%% file: main.tex
\documentclass[sigconf, screen]{acmart}
\AtBeginDocument{%
  \providecommand\BibTeX{{%
    \normalfont B\kern-0.5em{\scshape i\kern-0.25em b}\kern-0.8em\TeX}}}

\copyrightyear{2024}
\acmYear{2024}
\setcopyright{acmlicensed}\acmConference[KDD '24]{Proceedings of the 30th ACM SIGKDD Conference on Knowledge Discovery and Data Mining}{August 25--29, 2024}{Barcelona, Spain}
\acmBooktitle{Proceedings of the 30th ACM SIGKDD Conference on Knowledge Discovery and Data Mining (KDD '24), August 25--29, 2024, Barcelona, Spain}
\acmDOI{10.1145/3637528.3671676}
\acmISBN{979-8-4007-0490-1/24/08}
\usepackage[utf8]{inputenc} % allow utf-8 input
\usepackage[T1]{fontenc}    % use 8-bit T1 fonts
\usepackage{hyperref}       % hyperlinks
\usepackage{url}            % simple URL typesetting
\usepackage{amsfonts}       % blackboard math symbols
\usepackage{nicefrac}       % compact symbols for 1/2, etc.
\usepackage{microtype}      % microtypography
\usepackage{xcolor}         % colors

\usepackage{subcaption}

\usepackage{multirow}
\usepackage{algorithm}
\usepackage{algorithmic}
\usepackage{wrapfig}

\begin{document}

\newcommand{\fder}[2]{\frac{\partial#1}{\partial#2}}
\newcommand{\sder}[2]{\frac{\partial^2{#1}}{\partial{#2}^2}}
\newcommand{\kder}[2]{\frac{\partial^k{#1}}{\partial{#2}^k}}

\newcommand\dd[1]{\textcolor{blue}{[DD: #1]}}
\newcommand\ol[1]{\textcolor{red}{[OL: #1]}}
\newcommand\oa[1]{\textcolor{red}{[OA: #1]}}

\title{CONFIDE: Contextual Finite Difference Modelling of PDEs}

\author{Ori Linial}
\email{linial04@gmail.com}
\affiliation{%
  \institution{Technion – Israel Institute of Technology}
  \city{Haifa}
  \country{Israel}
}

\author{Orly Avner}
\affiliation{%
  \institution{Bosch Center for Artificial Intelligence}
  \city{Haifa}
  \country{Israel}}

\author{Dotan Di Castro}
\affiliation{%
  \institution{Bosch Center for Artificial Intelligence}
  \city{Haifa}
  \country{Israel}
}

\renewcommand{\shortauthors}{Ori Linial, Orly Avner \& Dotan Di Castro.}



\input{abstract}


\maketitle

\section{Introduction} \label{sec:intro} 
\input{introduction}

\section{Related Work} \label{sec:related}
\input{related}

\section{Method} \label{sec:method}
\input{method}

\section{Experiments} \label{sec:experiments}
\input{experiments}

\section{Conclusion} \label{sec:conclusions}
\input{conclusions.tex}

% \clearpage
% \newpage

\clearpage
\newpage

\bibliographystyle{ACM-Reference-Format}
\balance
\bibliography{main}

\clearpage
\newpage
\appendix
% \onecolumn

\input{appendix}

\end{document}

%% file: abstract.tex
\begin{abstract}
% We propose an explainable method for solving Partial Differential Equations by using a contextual scheme called \emph{PDExplain}. 
% During the training phase, our method is fed with data collected from a family of PDEs accompanied by the general form of this family. In the inference phase, a minimal sample collected from a phenomenon is provided, where the sample originates from the PDE family but not necessarily from the set of PDEs seen in the training phase. 
% Our algorithm can predict the PDE solution for future timesteps, while providing an explainable form of the PDE, a trait that can assist in data-driven modelling of phenomena in physical sciences. We include results of extensive experimentation, examining our method's quality both in terms of prediction error and explainability.
% We show how our algorithm can predict the PDE solution for future timesteps. Moreover, our method provides an explainable form of the PDE, a trait that can assist in modelling phenomena based on data in physical sciences. To verify our method, we conduct extensive experimentation, examining its quality both in terms of prediction error and explainability.
We introduce a method for inferring an explicit PDE from a data sample generated by  previously unseen dynamics, based on a learned context. 
The training phase integrates knowledge of the form of the equation with a differential scheme, while the inference phase yields a PDE that fits the data sample and enables both signal prediction and data explanation.
We include results of extensive experimentation, comparing our method to SOTA approaches, together with ablation studies that examine different flavors of our solution.
\end{abstract}

%% file: introduction.tex
Many scientific fields use the language of Partial Differential Equations (PDEs; \citealp{evans2010partial}) to describe the physical laws governing observed natural phenomena with spatio-temporal dynamics. 
Typically, a PDE system is derived from first principles and a mechanistic understanding of the problem after experimentation and data collection by domain experts of the field.
Well-known examples for such systems include Navier-Stokes and Burgers' equations in fluid dynamics, Maxwell's equations for electromagnetic theory, and Schr\"{o}dinger's equations for quantum mechanics.
Solving a PDE model could provide users with crucial information on how a signal evolves over time and space, and could be used for both prediction and control tasks.

% While creating PDE-based models holds great value, it is still a difficult task in many cases.
% For many complex real-world phenomena, we might only know some of the dynamics of the system.
% For example, an expert might tell us that a  heat equation PDE has a specific functional form
% %such as
% %\begin{align}
% %\label{eq:pde_coeffs}
% %        \fder{f}{t}=a(x,t,f) \frac{\partial^2 f }{\partial x^2}  + b(x,t,f)\frac{\partial f }{\partial x} + c(x,t,f),
% %\end{align}
% but we do not know the values of the diffusion and drift coefficient functions. We focus mainly on this case.
%In this case, the PDE system at hand is not solvable by standard methods.
Creating PDE-based models holds great value, but it is a difficult task in many cases. For some complex real-world phenomena, only some part of the systems dynamics is known, such as its structure, or functional form. For example, an expert might tell us that a  signal obeys the dynamics of a heat equation, without specifying the diffusion and drift coefficient functions. We focus mainly on this case, as explained in detail below.

The current process of solving PDEs over space and time is by using numerical differentiation and integration schemes. However, numerical methods may require significant computational resources, making the PDE solving task feasible only for low-complexity problems, e.g., a small number of equations.
An alternative common approach is finding simplified models that are based on certain assumptions and can roughly describe the problem's dynamics. A known example for such a model are the Reynolds-averaged Navier-Stokes equations \cite{reynolds1995dynamical}.
Building simplified models is considered a highly non-trivial task that requires special expertise, and might still not represent the phenomenon to a satisfactory accuracy.

In recent years, with the rise of Deep Learning (DL; \citealp{lecun2015deep}), novel methods for solving numerically-challenging PDEs were devised.
These methods have become especially useful thanks to the rapid development of sensors and computational power, enabling the collection of large amounts of multidimensional data related to a specific phenomenon.
In general, DL based approaches consume the observed data and learn a black-box model of the given problem that can then be used to provide predictions for the dynamics.
While this set of solutions has been shown to perform successfully on many tasks, it still suffers from two crucial drawbacks: (1) It offers no explainability as to why the predictions were made, and (2) it usually performs very poorly when extrapolating to unseen data. 

% \begin{wrapfigure}{r}{0.6\textwidth}
%   \begin{center}
%     \includegraphics[width=0.5\textwidth]{Figures/experiments/mech_dd.png}
%   \end{center}
%     \caption{Different approaches to PDE modeling, characterized by their ability to utilize knowledge regarding the underlying PDE.}
%     \label{fig:experiments.mech_dd}
% \end{wrapfigure}
In this paper, we offer a new hybrid modelling \citep{kurz2022hybrid} approach that can benefit from both worlds: it can use the vast amount of data collected on one hand, and utilize the partially known PDEs describing the observed natural phenomenon on the other hand. In addition, it can learn several contexts, thus employing the generalization capabilities of DL models and enabling zero-shot learning \citep{palatucci2009zero}.

Specifically, our model is given a general functional form of the PDE (i.e., which derivatives are used), consumes the observed data, and outputs the estimated coefficient functions. 
Then, we can then use off-the-shelf PDE solvers (e.g., PyPDE\footnote{\url{https://pypde.readthedocs.io/en/latest/}}) to solve and create predictions of the given task forward in time for any horizon.

Another key feature of our approach is that it consumes the spatio-temporal input signals required for training in an unsupervised manner, namely the coefficient functions that created the signals in the train set are unknown. This is achieved by combining an autoencoder architecture (AE; \citealp{kramer1991nonlinear,hinton2006reducing}) with a loss defined using the functional form of the PDE. As a result, large amounts of training data for our algorithm can be easily acquired.
Moreover, our ability to generalize to data corresponding to a PDE whose coefficients did not appear in the train set, enables the use of synthetic data for training.
Although our approach is intended to work when the PDE functional form is known, it is not limited to that scenario only.
In cases where we are given a misspecified model (when experts provide a surrogate model for instance), our model can eliminate some of the discrepancies using the extra function that is not a coefficient of one of the derivatives (the $p_0(x,t,u)$ function in  \eqref{eq:general_PDE}).

On the technical side, we chose to apply a finite difference approach in order to integrate the knowledge regarding the structure of the PDE family. This approach enables us to consume training data without requiring the corresponding boundary conditions.

A natural question for this setup is whether we are able to extract the ``correct'' coefficients for the PDE. The answer depends on the identifiability of the system, a trait that does not hold for many practical scenarios. We therefore focus on finding the coefficients that best explain the data, making prediction of the signal forward in time possible. Practitioners will find the estimated coefficients useful even if they are not exact, since they may convey the shape, or dynamics, of unknown phenomena.

Our motivation comes from the world of electric vehicle batteries, where PDEs are used to model battery charging, discharging and aging. For a specific type of battery, the set of equations has a common form, with different coefficients for each battery instance. The data describing battery dynamics is gathered by battery management systems in the vehicle, and also in the lab. It is then used to calibrate the equation-based model, in order to later generate predictions and analyze battery behavior. Traditional techniques for model calibration are based on direct optimization, and suffer from two drawbacks: (1) they are extremely time consuming, (2) they do not leverage data from one battery in the dataset to another. Our approach solves both issues: model calibration is achieved by inference rather than optimization, and the learned context facilitates transfer of knowledge between batteries. The first improvement is straightforward, and the second one stems from the context-based architecture we introduce. This architecture enables us to estimate the coefficients of a given battery based on a smaller amount of data when compared to the traditional approach.

We summarize our contribution as follows:
\begin{enumerate}
    \item Harnessing the information contained in large datasets belonging to a phenomenon which is related to a PDE functional family in an unsupervised manner. Specifically, we propose a regression based method, combined with a finite difference approach.
    \item Proposing a DL encoding scheme for the context conveyed in such datasets, enabling generalization for prediction of unseen samples based on minimal input, similarly to zero-shot learning.
    \item Extensive experimentation with the proposed scheme, examining the effect of context and train set size, along with a comparison to different previous methods.
\end{enumerate}

The paper is organized as follows. In Section \ref{sec:related} we review related work. In Section \ref{sec:method} we present the proposed method and in Section \ref{sec:experiments} we provide experiments to support our method. Section \ref{sec:conclusions} completes the paper with conclusions and future directions.

%% file: related.tex
Creating a neural-network based model for approximating the solution of a PDE has been studied extensively over the years, and dates back more than two decades \citep{lagaris1998artificial}.
We divide deep learning based approaches by their ability to incorporate mechanistic knowledge in their models, and by the type of information that can be extracted from using them.
Another distinction between different approaches is their ability to handle datasets originating from different contexts.
From a PDE perspective, a different context could refer to having data signals generated with different coefficients functions ($p_l$ in \eqref{eq:general_PDE}).
In many real-world applications, obtaining observed datasets originating from a single context is impractical.
For example, in cardiac electrophysiology \citep{neic2017efficient}, patients differ in cardiac parameters like resistance and capacitance, thus representing different contexts. 
In fluid dynamics, the topography of the underwater terrain (bathymetry) differs from one sample to another \citep{hajduk2020bathymetry}.

The first line of work is purely data-driven methods.
These models come in handy when we observe a spatio-temporal phenomenon, but either don't have enough knowledge of the underlying PDE dynamics, or the known equations are too complicated to solve numerically (as explained thoroughly by \citet{wang2021physics}). 
Recent advances demonstrate successful prediction results that are fast to compute (compared to numerically solving a PDE), and also provide decent predictions even for PDEs with very high dimensions \citep{brandstetter2022message,li2020fourier,yin2022continuous,gupta2022towards,han2018solving,lu2019deeponet,pfaff2020learning}. 
However, the downside of these approaches is not being able to infer the PDE coefficients, which may hold valuable information and explanations as to why the model formed its predictions.

\begin{figure*}[t]
     \centering
     \begin{subfigure}[b]{0.45\textwidth}
         \centering
         \includegraphics[width=\textwidth]{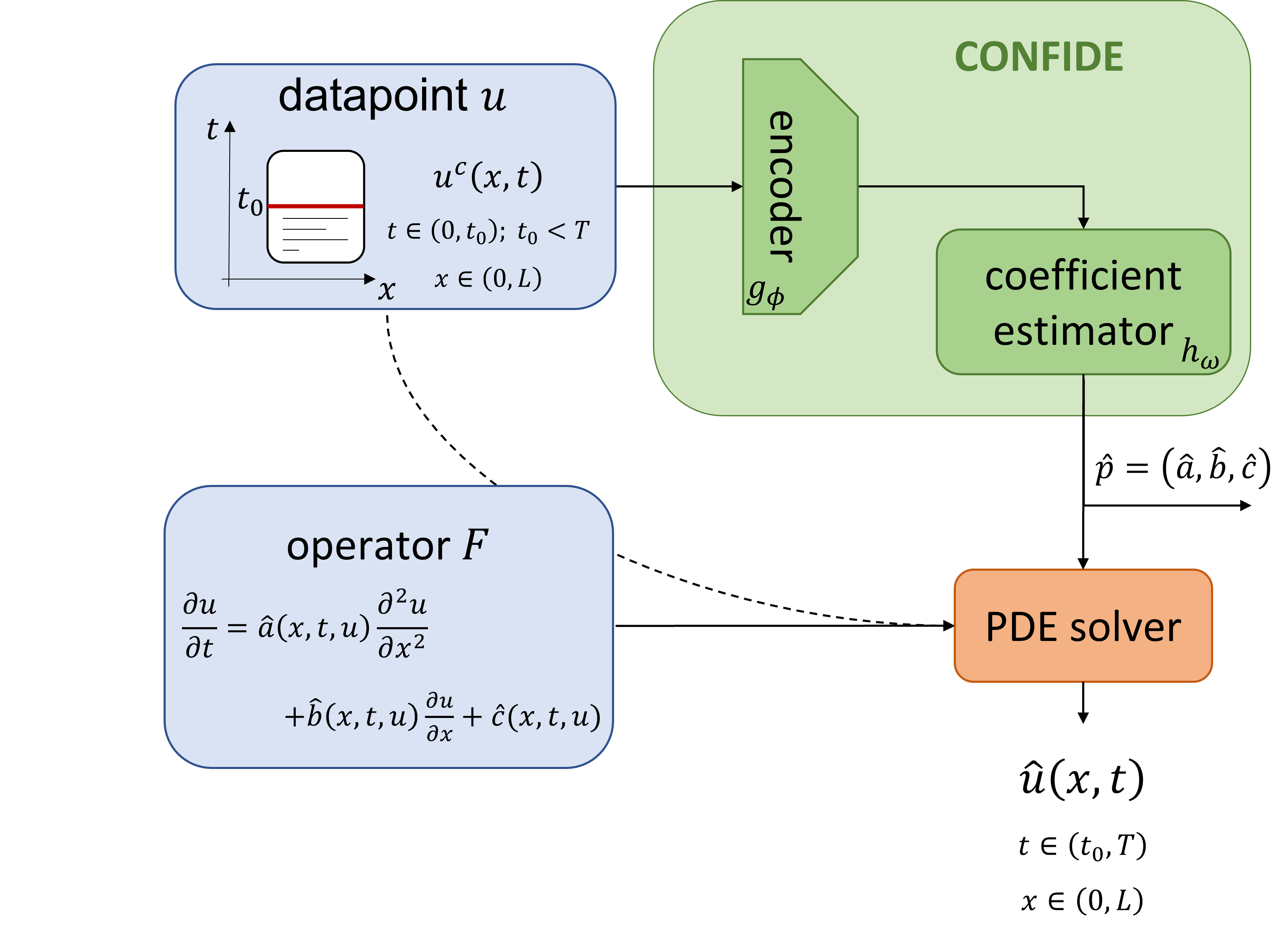}
         \caption{}
         \label{fig:method.inference}
     \end{subfigure}
     \hfill
     \begin{subfigure}[b]{0.45\textwidth}
         \centering
         \includegraphics[width=\textwidth]{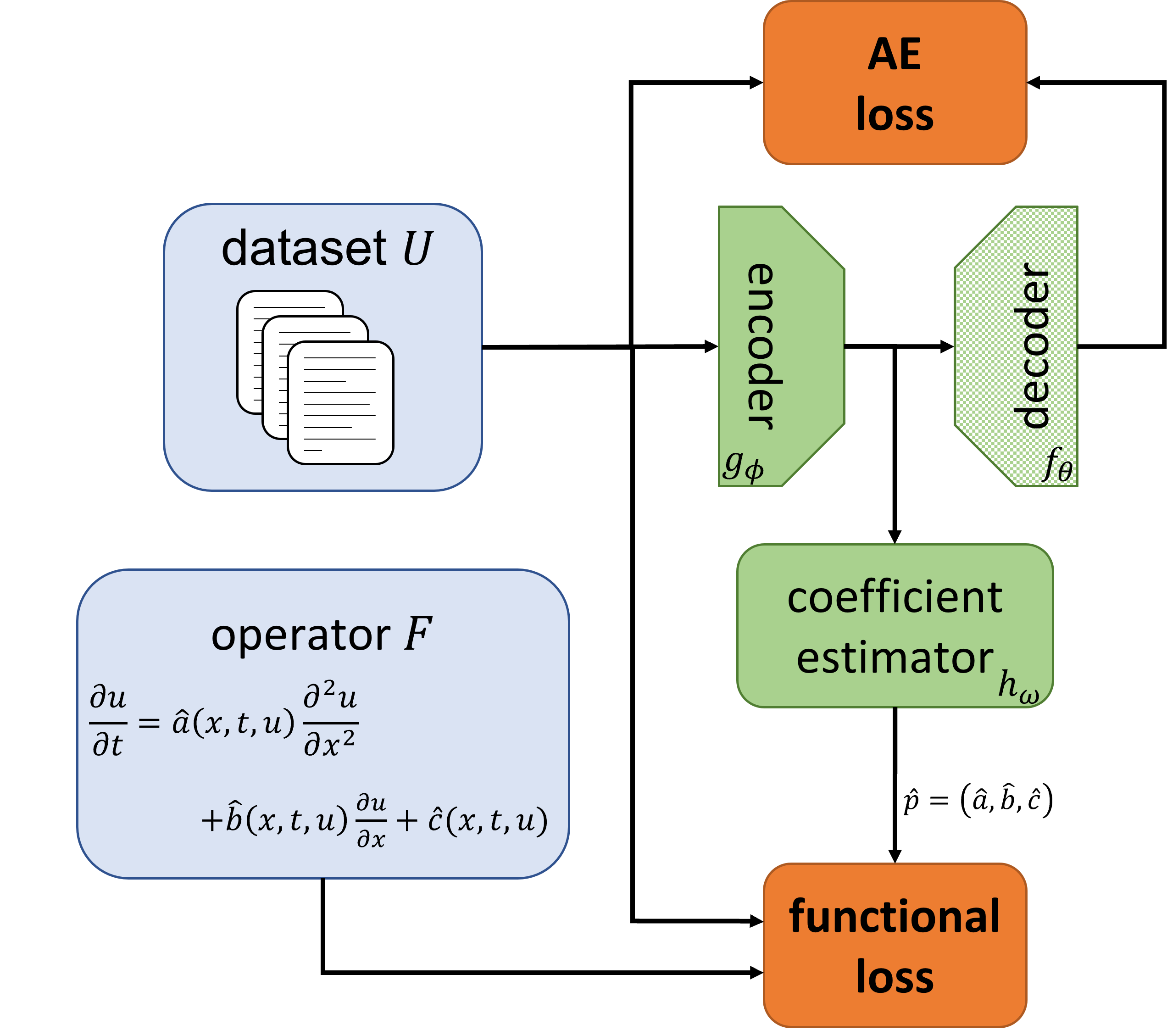}
         \caption{}
         \label{fig:method.training}
     \end{subfigure}
     \hfill
    \caption{(a) Inference process: given an observed spatio-temporal signal, CONFIDE estimates the PDE coefficients that best describe it. These can be plugged into a PDE solver together with the known operator form $F$, and an initial condition (dashed line) to obtain a prediction of the signal for future time-steps. (b) Training process: 
    In each iteration, CONFIDE observes a set of signals generated by the same family of PDEs.
    For each train signal, CONFIDE evaluates the PDE coefficients best describing the observed signal, and all the spatio-temporal derivatives that are known to be in the functional form of the PDE (e.g., $\fder{u}{t}$, $\sder{u}{x}$, ...). The derivatives and coefficients are then plugged into the operator $F$ which is then used to minimize the functional loss (as in Eq.~\eqref{eq:coefficient_estimator_loss}) and train a context-based coefficient estimator.}
    \Description{The training and inference process of the CONFIDE algorithm. Inference process: given an observed spatio-temporal signal, CONFIDE estimates the PDE coefficients that best describe it. These can be plugged into a PDE solver together with the known PDE form, and an initial condition to obtain a prediction of the signal for future time-steps. Training process: 
    In each iteration, CONFIDE observes a set of signals generated by the same family of PDEs.
    For each train signal, CONFIDE evaluates the PDE coefficients best describing the observed signal, and all the spatio-temporal derivatives that are known to be in the functional form of the PDE. The derivatives and coefficients are then plugged into the PDE which is then used to minimize the functional loss and train a context-based coefficient estimator.}
    \label{fig:three graphs}
\end{figure*}

The second type of data-driven methods are approaches that utilize PDE forms known beforehand to some extent.
Works that adopt this approach can usually utilize the given mechanistic knowledge and provide reliable predictions, ability to generalize to unseen data, and in some cases even reveal part of the underlying PDE coefficient functions.
However, their main limitation is that they assume the entire training dataset is generated by a single coefficient function and only differ in the initial conditions (or possibly boundary conditions).
PDE-NET \citep{long2018pde}, its followup PDE-NET2 \citep{long2019pde}, DISCOVER \citep{du2022discover}, PINO \citep{li2021physics} and sparse-optimization methods \citep{schaeffer2017learning,rudy2017data} (expanding the idea originally presented on ODEs \cite{brunton2016discovering,champion2019data}), are not given the PDE system, but instead aim to learn some representation of the underlying PDE as a linear combination of base functions and derivatives of the PDE state.
PINN \citep{raissi2019physics} and NeuralPDE \citep{zubov2021neuralpde} assume full knowledge of the underlying PDE including its coefficients, and aim to replace the numerical PDE solver by a fast and reliable model. 
They also provide a scheme for finding the PDE parameters as scalars, but assume the entire dataset is generated by a single coefficient value, while we assume each sample is generated with different coefficient values which could be functions of time, space and state (as described in \eqref{eq:general_PDE}). In \citet{negiar2022learning}, the authors incorporate knowledge of the PDE structure as a hard constraint while learning to predict the solution to the PDE.
Similarly, Learning-informed PDEs \citep{dong2022optimization,aarset2022learning} suggest a method that assumes full knowledge of the PDE derivatives and their coefficient functions, and infers the free coefficient function (namely $p_0(x,t,u)$ in \eqref{eq:general_PDE}). 
In \citep{lim2022physics}, the authors apply a finite difference approach to PINNs. Another approach for learning the solution to PDEs, that also uses a neural representation, is introduced in \citep{chen2022crom}. Recently, in \citep{subramanian2023towards}, the authors have suggested a Foundation Model framework for predicting solutions of PDEs.

The last line of work, and closer in spirit to ours, includes context-aware methods that assume some mechanistic knowledge, with each sample in the train set generated by different PDE coefficients (we also refer to this concept as having different context) and initial conditions.
CoDA \citep{kirchmeyer2022generalizing} provides the ability to form predictions of signals with unseen contexts, but does not directly identify the PDE parameters.
GOKU \citep{linial2021generative} and ALPS \citep{yanglearning} provide context-aware inference of signals with ODE dynamics, when the observed signals are not the ODE variables directly.
Another important paper introduces the APHYNITY algorithm \citep{yin2021augmenting}, which also presents an approach to inferring PDE parameters from data. This work handles the scenario of fixed coefficients, as opposed to our ability to handle coefficients that are functions. Also, the case of coefficients that differ between samples is addressed only briefly, with a fixed, rather high, context ratio.

In Section \ref{sec:experiments} we present comparisons to several carefully selected baselines mentioned above. 
The first approach is Neural-ODE \citep{chen2018neural} (also referred as Latent-ODE in its follow-up paper \citep{rubanova2019latent} when prompted to solve a time-series prediction task). 
% While the main focus of the Neural-ODE is overcoming the challenge of backpropagation through an ODE solver, it also addresses predicting an arbitrary time series that does not necessarily follow a known ODE. 
The Neural-ODE approach infers the initial conditions of an arbitrary latent trajectory from some high dimension, integrate it through time by assuming it follows a learnable dynamics function, and outputs the future predictions of the observed signal.
Specifically, a given PDE could be considered as a time series from high dimension, and Neural-ODE should be able to learn its dynamics, and provide future predictions.
The second and third approaches are Fourier-Neural-Operators (FNO) \citep{li2020fourier} and UNet \citep{gupta2022towards} which are designed to provide a faster alternative to solving a PDE using a classic PDE-solver.
Both approaches expand the Deep-Operator-Network (DeepONet) \citep{lu2019deeponet} work, where a the model learns a mappings between spaces of functions. 
In the PDE prediction task, one example is that they learn the connection between the PDE initial conditions and the solution at some required time $t=\tau$. 
% In \citep{li2020fourier}, the authors introduce the Fourier Neural Operator (FNO), a deep learning architecture that learns mappings between spaces of functions, applying it to predicting solutions of PDEs of a given structure. Another recent approach, proposed in \citep{gupta2022towards}, applies the well-known U-Net architecture to PDE modelling with different parameter values and temporal resolutions. 
The last baseline we compare to is DINo \citep{yin2022continuous}. 
In this approach, the model applies a scheme combining Neural-ODEs with Implicit Neural Representations, which are implemented as FourierNets, to PDE forecasting tasks.

%% file: method.tex
The data we handle is a set of spatio-temporal signals generated by an underlying PDE, only the form of which is known. The coefficient functions determining the exact PDE are unknown and may be different for each collection of data. 
Our goal is to estimate these coefficient functions and provide reliable predictions of the future time steps of the observed phenomenon.
The proposed method comprises three subsequent parts: (1) Creating a compact representation of the given signal, (2) estimating the PDE coefficients, and (3) solving the PDE using the acquired knowledge. For ease of exposition we focus on parabolic PDEs in this section, however the extension to other types of PDEs is straightforward.

\subsection{Problem Formulation} 
We now define the problem formally.
Let $U(x,t)$ denote a spatio-temporal function defined over some compact spatial domain $\Omega \subseteq \mathbb{R}^d$, where $d$ is the number of spatial variables, and a temporal domain $\mathbb{R}$. $U(x,t)$ maps between points in the spatial domain $x\in\Omega$ at some point in time $t$ to an $n$-dim vector, where $n$ is the number of observed variables.
In other words, $U(x,t):\Omega\times\mathbb{R}\rightarrow\mathbb{R}^n$. 
In addition, the initial value of the function $U(x,t=0)$ changes between observed signals, therefore sampled from some unknown probability function $U(x,t=0)\sim P_{u_0}$.
An observed signal $u(x,t)$ is therefore a projection of the function $U(x,t)$ on a finite discrete observation grid and on discrete times.
In our formulation, we make the problem harder by considering the case where each signal $u(x,t)$ could also originate from a PDE with different coefficients, thus differing between observed signals.
This means that unlike most related works, in this work we assume that every observed signal corresponds to a different instance of the PDE family.
We refer to the signals with different coefficients as $u^c(x,t)$, where $c$ stands for \emph{context}, which changes between observed signals.
For example, we might know that $u^c(x,t)$ follows the Navier-Stokes equations, but some coefficients might change between observations (like the viscosity of the fluid).

Generally, $u^c(x,t)$ could obey any PDE, but in this work we focus on parabolic PDEs, hence the signal $u^c(x,t)$ is the solution of a $k$-th order PDE of the general form
\begin{align} \label{eq:general_PDE}
    \fder{u}{t}= \sum_{l=1}^k p_l(x,t,u) \frac{\partial u^l}{\partial x^l}+ p_0(x,t,u),
\end{align}
with a vector of coefficient functions $p=(p_0, \ldots,p_k)$. We adopt the notation of \citet{wang2021physics} and refer to a family of PDEs characterized by a vector $p$ as an operator $F(p, u)$, where solving $F(p,u)=0$ yields solutions of the PDE.

The problem we solve is as follows: given an observed signal $u^c(x,t)$, at times $t=0,\ldots,t_0$ that solves a PDE of a \emph{known} operator $F$ with an \emph{unknown} coefficient vector $p$, we would like to (a) estimate the coefficient vector $\hat{p}$ and (b) predict the signal at future times $t=t_0,\ldots,T$, for some $T>t_0$.

Our solution is a concatenation of two neural networks, which we call \emph{CONFIDE}. Its input is an observed signal $u^c(x,t=0,\ldots,t_0)$, and its output is a vector $\hat{p}$. We feed this vector into an off-the-shelf PDE solver together with the operator $F(p,u)$ to obtain the predicted signal $\hat{u}(x,t=t_0,\ldots,T)$. An explanation of our numerical scheme appears in Section \ref{sec:app_num}.

\subsection{CONFIDE Inference} 

\label{ssec:method.PDExplain_inference}
We begin by outlining our inference process, presented in Fig.~\ref{fig:method.inference}.
The input to this process is an observed signal $u^{c}(x,t)$, defined on some compact spatial domain $\Omega$, for times $t\in[0,t_0]$ and an operator $F$ 
% (e.g., the one introduced in \eqref{eq:general_PDE} for $k=2$). 
The input is fed into the CONFIDE component, which generates the estimated coefficients $\hat{p}$.
For example, taking $k=2$ in the example in \eqref{eq:general_PDE} results in a coefficient vector $\hat{p}=(\hat{a},\hat{b},\hat{c})$. 
The PDE solver then uses this estimate to predict the complete signal, $\hat{u}(x,t)$, $x\in\Omega$, $t\in[t_0,T]$. 
An important feature of our approach is the explicit prediction of the coefficient functions, which contributes to the explainability of the solution.

The observation $u^{c}(x,t=t_0)$ serves as an initial condition for the prediction and also represents the dynamics of the signal for estimating the PDE coefficients. In the sequel we refer to it as ``context''. The ratio of between the observed times and the required prediction time is denoted by $\rho$, such that $t_0=\rho T$, and is a hyper-parameter of our algorithm.
% \begin{figure}[ht]
%     \centering
%     \includegraphics[width=0.9\columnwidth]{Figures/inference.png}
%     \caption{Inference process. Dashed line is initial condition, supplied to the PDE solver together with the estimated coefficients and operator $F$.}\label{fig:method.inference}
% \end{figure}

\begin{algorithm}[htb]
   \caption{CONFIDE inference scheme}
   \label{alg:algorithm1}
\begin{algorithmic}
    \STATE {\bfseries Input:} observation $u^{c}(x,t) $,  operator $F$, trained networks: encoder $g_{\phi}$, coefficient estimator $h_{\omega}$
    \STATE $\hat{p} \gets h_{\omega}(g_{\phi}(u^c))$
    \STATE $\hat{u} \gets \textrm{PDE\_solve}(F, \hat{p}, u^c(x,t=t_0))$
    \STATE return $\hat{u}, \hat{p}$
\end{algorithmic}
\end{algorithm}
\begin{algorithm}[htb]
   \caption{Algorithm for training CONFIDE}
   \label{alg:algorithm2}
\begin{algorithmic}
   \STATE {\bfseries Input:} dataset $\mathcal{D}$, operator $F$, time $t_0$, loss weight $\alpha$, number of epochs $N_{e}$
   \STATE {\bfseries Init:} random weights in encoder $g_{\phi}$, decoder $f_\theta$, coefficient estimator $h_{\omega}$
   \FOR{epoch in $N_{e}$}
    \STATE $\mathcal{L} \gets 0$
    \STATE $\{u_i^c\}_{i=1}^N \gets$ {Random batch of $N$ observations from $\mathcal{D}$}
    \FOR{$u_i^c$ in batch}
        \STATE $\hat{p}_i\gets h_{\omega}(g_{\phi}(u_i^c))$
        \STATE $\mathcal{L}_{\textrm{AE}} \gets (u_i^c-f_\theta(g_{\phi}(u_i^c), u_i(t=0)))^2$
        \STATE $\tau \gets$ Random value from $[0,t_0]$      \STATE $\mathcal{L}_{\textrm{coef}} \gets \left\Vert F(\hat{p}_i,u_i^c(t=\tau))\right\Vert^2$
        \STATE $\mathcal{L} \gets \mathcal{L} + \alpha \cdot \mathcal{L}_{\textrm{AE}} + (1-\alpha) \cdot \mathcal{L}_{\textrm{coef}}$
    \ENDFOR
    \STATE $\phi, \theta, \omega \gets \mathop{\arg \min \mathcal{L}}$
   \ENDFOR
\end{algorithmic}
\end{algorithm}

\subsection{CONFIDE Training} \label{ssec:method.PDExplain_training}
The training process is presented in Fig.~\ref{fig:method.training}. Its input is a dataset $\mathcal{D}$ that consists of $N$ signals $\{u^c_i(x,t)\}_{i=1}^N$, which are solutions of $N$ PDEs that share an operator $F$ but have unique coefficient vectors $\{p_i\}_{i=1}^N$. We stress that the vectors $p_i$ are unknown even at train time. 
The signals are defined on some domain $\Omega$, and for times $t\in[0,t_0]$.
% The support of the signals is $x\in[0,L]$, $t\in[0,T]$. 
The loss we minimize is a weighted sum of two components: (i) the autoencoder reconstruction loss 
% (AE; \citealp{hinton2006reducing})
, which is defined in \eqref{eq:autoencoder_loss}, and (ii) the functional loss as defined in  \eqref{eq:coefficient_estimator_loss}. 

\begin{figure*}[t]
     \centering
     \begin{subfigure}{0.48\textwidth}
         \centering
         \includegraphics[width=\textwidth]{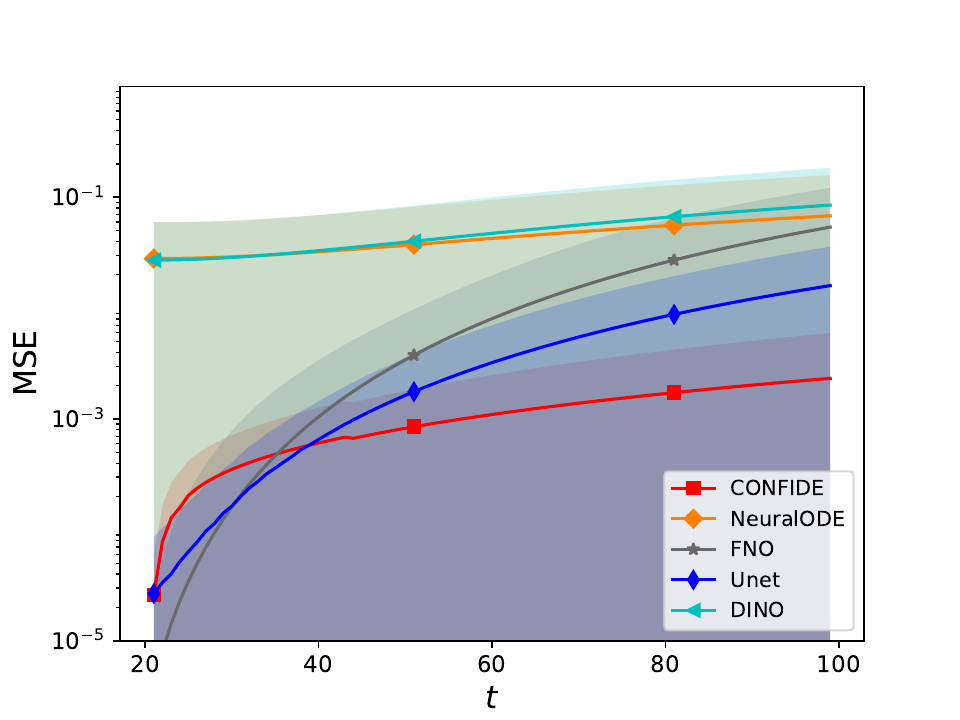}
         \caption{}
         \label{fig:experiments.pred_rollout}
     \end{subfigure}
     % \hfill
     \begin{subfigure}{0.48\textwidth}
         \centering
         \includegraphics[width=\textwidth]{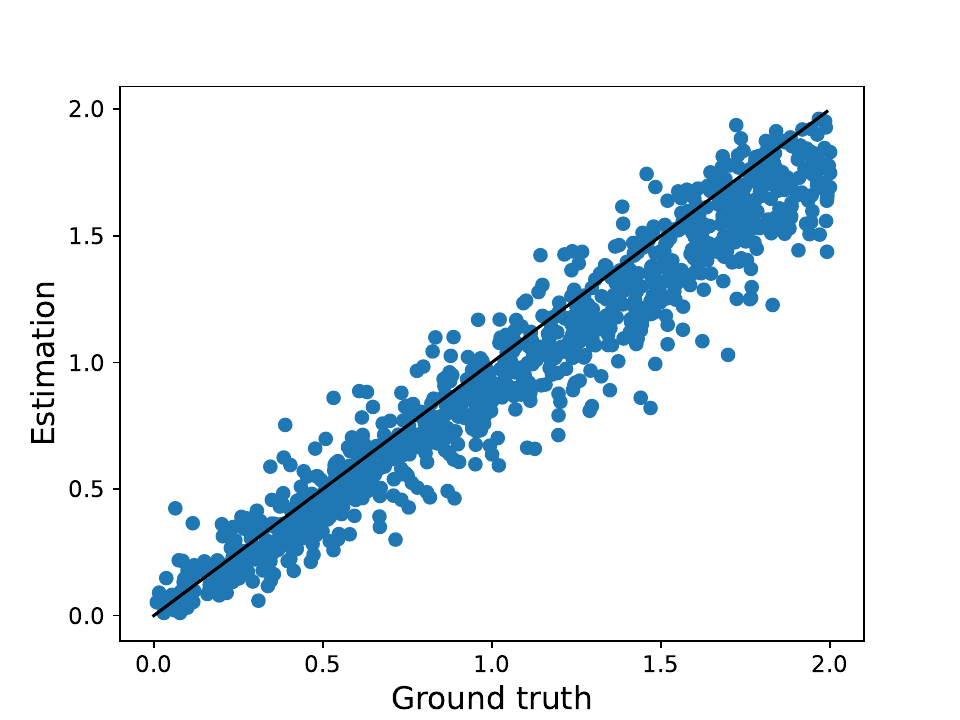}
         \caption{}
         \label{fig:experiments.param_a_r2}
     \end{subfigure}
     \hfill
    \caption{Constant coefficients (Section~\ref{ssec:pde_exp1}). \textbf{(a)} Prediction error vs. prediction horizon, for different algorithms. CONFIDE, in red, is our approach. \textbf{(b)} Estimated value of the $\partial^2 u / \partial x^2 $ coefficient vs. ground truth, for test set ($R^2 = 0.93$).}
    \Description{Results of the PDE with Constant coefficients. of the left, Prediction error vs. prediction horizon, for different algorithms where CONFIDE beats all other algorithms. on the right, example of the estimated value of one of the PDE coefficients  vs. ground truth, for test set}
\end{figure*}

CONFIDE comprises two parts: (1) an encoder and (2) a coefficient estimator. The encoder's goal is to capture the dynamics driving the signal $u_i$, thus creating a compact representation for the coefficient estimator. The encoder is trained on signals $u^c_i$ in the train set. Each signal is of size $t_0 \times$ amount of spatial points (e.g., for $\Omega = [0,L]$, the size is  $t_0 \cdot L$ points). 

% \begin{figure}[ht]
%     \centering
%     \includegraphics[width=0.9\columnwidth]{Figures/training.png}
%     \caption{Training process.}\label{fig:method.training}
% \end{figure}

The encoder loss is the standard AE reconstruction loss, namely the objective is
\begin{align}
\label{eq:autoencoder_loss}
\mathop{\min}\limits_{\theta,\phi}\mathcal{L_{\textrm{AE}}} = 
\mathop{\min}\limits_{\theta,\phi}\sum_{i=1}^N \textrm{loss}(u_i^c - f_{\theta}(g_{\phi}(u_i^c))),
\end{align}
where $f_{\theta}$ is the decoder, $g_{\phi}$ is the encoder and $\textrm{loss}(\cdot,\cdot)$ is a standard loss function (e.g., $L^2$ loss).

The second component is the coefficient estimator, whose input is the encoded context and the signal  The estimated coefficients output by this component, together with the operator $F$, and the signal at at some random time $t=\tau\in[0,t_0]$, form the functional objective:
\begin{align}
\label{eq:coefficient_estimator_loss}
\mathop{\min}\limits_{\omega}\mathcal{L_{\textrm{coef}}} = 
    \mathop{\min}\limits_{\omega}\sum_{i=1}^N 
    \left\Vert F(\hat{p}_\omega,u^c_i(x,\tau)\right\Vert^2,
\end{align}
where $\omega$ represents the parameters of the coefficient estimator network, and $\hat{p}$ is the estimator of $p$ at time $\tau$, acquired by applying the network $h_\omega$ to the output of the encoder.
This design enables CONFIDE to learn a parameter vector which can depend on time, space, and the observation $u$.

The two components are trained simultaneously, and the total loss is a weighted sum of the losses in \eqref{eq:autoencoder_loss} and \eqref{eq:coefficient_estimator_loss}: 
$\mathcal{L} = \alpha \cdot \mathcal{L}_{\textrm{AE}} + (1-\alpha) \cdot \mathcal{L}_{\textrm{coef}}$, where $\alpha\in(0,1)$ is a hyper-parameter.

\paragraph{\textbf{Initial-conditions aware autoencoder}}
To further aid our model in learning the underlying dynamics of the observed phenomenon, we include the observed initial conditions of the signal (i.e., $u_i(t=0)$) along with the latent context vector (i.e., $g_{\phi}(u^c_i)$) as input to the decoder network.
This modification enables the model to learn a context vector that better represents the dynamics of the phenomenon, rather than other information such as the actual values of the signal.

We experimented with removing the decoder and training the networks using the functional loss alone, and without including the initial conditions as an input to the decoder.
In both cases, results proved to be inferior, suggesting that the autoencoder loss helps the model to focus on the underlying dynamics of the observed signal.

To summarize this section, we present the inference scheme in Algorithm~\ref{alg:algorithm1}, and the full training algorithm  in Algorithm~\ref{alg:algorithm2}.

%% file: experiments.tex
% We devote this section to two types of analyses: (a) a comparison of our approach to other solutions, and (b) an analysis of our approach in different scenarios.

% The dataset we use includes $10,000$ samples of size $100\times 40[t\: \textrm{points} \times x\: \textrm{points}]$, 

We devote this section to analyse and compare our approach to other solutions, on four different systems of PDEs: (1) constant coefficients, (2) Burgers' equations, (3) 2D-FitzHugh-Nagumo, and (4) 2D-Navier-Stokes equation.
For each PDE task, we created a dataset of signals generated from a PDE with different coefficients.
We could not use off-the-shelf datasets, such as those appearing in PDEBench \citep{takamoto2022pdebench}, since each of the datasets there is generated from a single constant function (i.e., all data samples have the same context). We used well-known equations, therefore our datasets can serve as a benchmark for the emerging field of contextual PDE modelling. 
We stress the fact that the test set contains signals generated by PDEs with coefficient vectors that \emph{do not} appear in the training data, hence demonstrating different dynamics than the ones the model observed during training. 
In that sense, the task at hand is a zero-shot prediction problem. 
More information about dataset creation can be found in the appendix.

% \begin{figure}[h!]
%     \centering
%     \includegraphics[width=\columnwidth]{Figures/experiments/burger_demonstration_single_plots_621.pdf}
%     \caption{A solution of the Burgers' equation. The black plot in each figure displays the ground truth. Rows correspond with the predicted solution by the respective algorithm (top row for CONFIDE displayed in red). Each column shows the solution at a different time point. The rightmost column shows the solution at $t=100$ zoomed to demonstrate the differences.}\label{fig:experiments.burgers_demo}
% \end{figure}

We benchmark the performance of CONFIDE against several state of the art approaches:
\begin{enumerate}
    \item Neural ODE, based on the algorithm suggested by \citet{chen2018neural}, Section 5.1 (namely, Latent ODE).
    \item Fourier Neural Operator (FNO), introduced by \citet{li2020fourier}.
    \item U-Net, as presented by \citet{gupta2022towards}.
    \item DINo, as presented by \citet{yin2022continuous}.
\end{enumerate}
Additional details regarding the implementation of baselines can be found in Section ~\ref{app:implementation}.

\begin{figure*}[t]
     \centering
     \begin{subfigure}[b]{0.49\textwidth}
         \centering
         \includegraphics[width=\columnwidth]{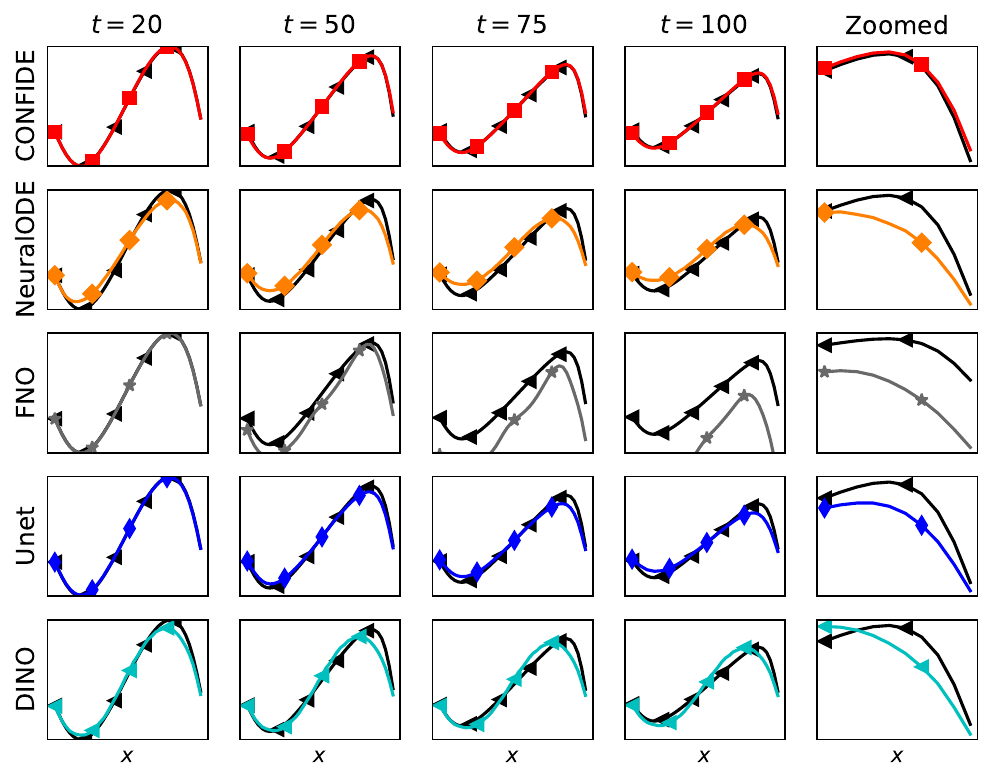}
         \caption{}
        \label{fig:experiments.burger_demo}
     \end{subfigure}
     % \hfill
     \begin{subfigure}[b]{0.49\textwidth}
         \centering
         \includegraphics[width=\textwidth]{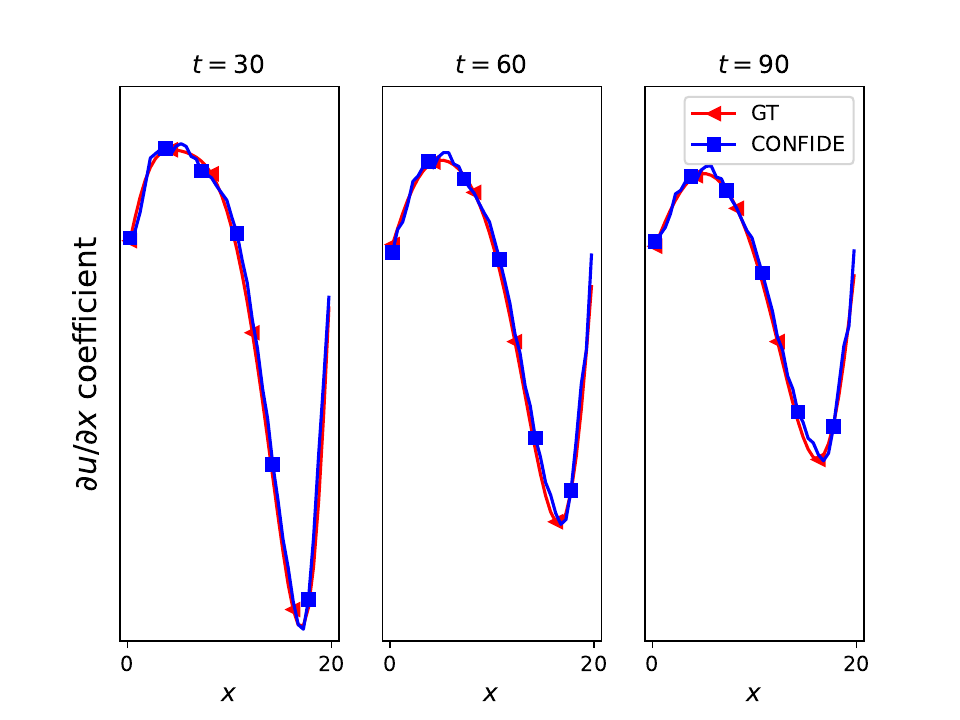}
         \caption{}
         \label{fig:experiments.burger_params_b}
     \end{subfigure}
     \hfill
    \caption{Burgers' PDE:
    \textbf{(a)} A solution of the Burgers’ equation. The black plot in each figure displays the ground truth. Rows correspond with the predicted solution by the respective algorithm (top row for CONFIDE displayed in red). Each column shows the solution at a different time point. The rightmost column shows the solution at t = 100 zoomed to demonstrate the differences.
    \textbf{(b)} Estimation of the coefficient function $b(x,t,u)$ of the Burgers' equation from \eqref{eq:pde_exp2}. CONFIDE manages to accurately estimate the spatio-temporal dynamics of the coefficient, based on a context ratio of $\rho=0.2$.}
    \Description{Results on Burgers' PDE. on the left an example, how it evolves over time, and how CONFIDE is closest to the ground truth. On the right, an example of the coefficient function inferece.}
\end{figure*}

\subsection{Second Order PDE with Constant Coefficients}\label{ssec:pde_exp1}
The first family of PDEs used for our experiments is: 
\begin{align}
\label{eq:pde_exp1}
        \fder{u}{t}=a \sder{u}{x} + b \fder{u}{x} + c,
\end{align}
where $p=(a,b,c)$ are constants but differ between signals. 
Figure~\ref{fig:experiments.pred_rollout} demonstrates the clear advantage of our approach, which increases with the prediction horizon (note the logarithmic scale of the vertical axis, representing the MSE of prediction). Since CONFIDE harnesses both mechanistic knowledge and training data, it is able to predict the signal $\hat{u}(x,t)$ several timesteps ahead, while keeping the error to a minimum.

Another result for this set of experiments appears in Figure~\ref{fig:experiments.param_a_r2}. Here, we plot the estimated value of parameter $a$ of \eqref{eq:pde_exp1}, against its true value. The plot and the high value of $R^2$ demonstrate the low variance of our prediction, with a strong concentration of values along the $y=x$ line.

Section~\ref{sec:app.ablation} presents the results of an ablation study on the hyper-parameters of CONFIDE for this equation.

\subsection{Burgers' equation}
Another family of PDEs we experiment with is the quasi-linear Burgers' equation, whose general form is
\begin{align}
    \label{eq:pde_exp2}
        \fder{u}{t}=a\sder{u}{x} + b(u)\fder{u}{x},
\end{align}
where $b(x,t,u)=-u$, as presented in \cite{bateman1915some}. We note that this equation is quasi-linear since its drift coefficient $b(x,t,u)$ depends on the solution $u$ itself. The dataset for our experiments consists of 10000 signals with different values of $a$ and the same $b(u)=-u$, both unknown to the algorithm a priori.
We begin with a demonstration of a signal $u(x,t)$ and its prediction $\hat{u}(x,t)$ in Figure~\ref{fig:experiments.burger_demo}. As can be seen both visually and from the value of the MSE (in each panel's title), our approach yields a prediction that stays closest to the ground truth (GT), even as the prediction horizon (vertical axis) increases.

\begin{figure}[t]
    \centering
    \includegraphics[width=\columnwidth]{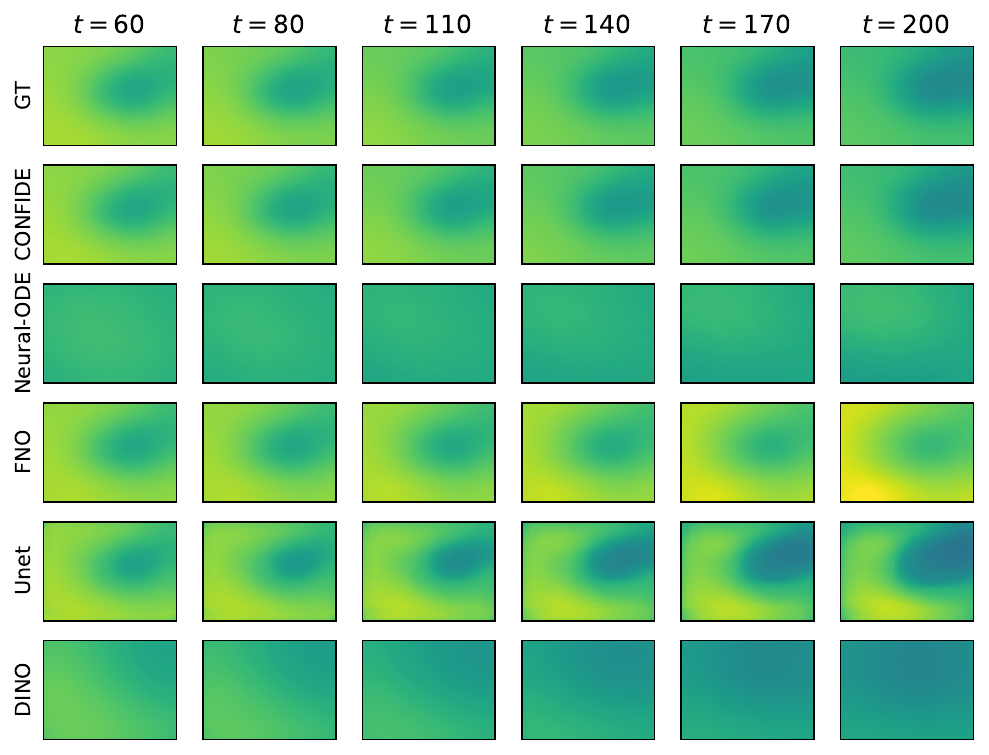}
    \caption{2D-FitzHugh-Nagumo PDE: Figures in the top row show the ground truth of $R_v$ for different time points, and the rows below show the estimation of it by the different approaches.
    CONFIDE estimates $R_v$ directly and near-perfectly recovers the unknown part of the PDE even as the prediction horizon increases. For the other algorithms we evaluated $R_v=u-v$ from the predictions of $u$ and $v$.}\label{fig:experiments.fn2d}
    \Description{Results on predicting the results of FitzHugh-Nagumo PDE, where CONFIDE is closest to the ground truth.}
\end{figure}

In Figure~\ref{fig:experiments.burger_params_b} we focus on the ability to accurately predict coefficient functions with spatio-temporal dynamics, in this case: the coefficient $b(x,t,u)$ of \eqref{eq:pde_exp2}. The panels correspond to different points in time, showing that the coefficient estimator tracks the temporal evolution successfully.

\subsection{FitzHugh-Nagumo equations}
The next family of PDEs we examine is the FitzHugh-Nagumo PDE \citep{klaasen1984stationary} consisting of two equations:
\begin{align}
    \label{eq:fn2d_eq}
        \fder{u}{t} = a \Delta u + R_u(u,k,v), \qquad
        \fder{v}{t} = b \Delta v + R_v(u,v),
\end{align}
where $a$ and $b$ represent the diffusion coefficients of $u$ and $v$, and $\Delta$ is the Laplace operator.
For the local reaction terms, we follow \citet{yin2021augmenting} and set $R_u(u,k,v)=u-u^3-k-v$, and $R_v(u,v)=u-v$. The PDE state is $(u,v)$, defined on the 2-D rectangular domain $(x,y)$ with periodic Neumann boundary conditions.

\begin{figure}[t!]
\centering
\includegraphics[width=\columnwidth]{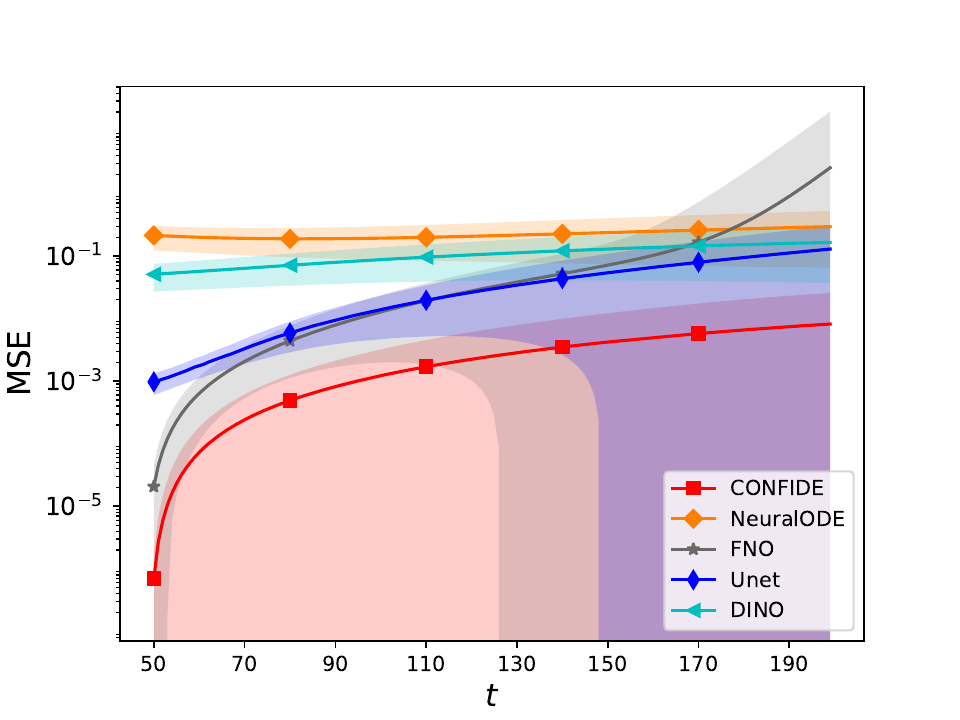}
\caption{2D-FitzHugh-Nagumo PDE: prediction error as horizon increases, for different approaches.}
\label{fig:experiments.FN2D_error}
\Description{Prediction results on FitzHugh-Nagumo PDEs.}
\end{figure}

The dataset created for this task consists of 1000 signals, each with a different value of $k$. 
We compare the prediction generated by CONFIDE to those yielded by other approaches, and present a typical result in Fig.~\ref{fig:experiments.fn2d}. In Fig.~\ref{fig:experiments.FN2D_error} we present the prediction error as a function of the prediction horizon, once again comparing CONFIDE to the baselines.

\subsection{Navier-Stokes equation}
\begin{figure}[t!]
    \centering
    \includegraphics[width=\columnwidth]{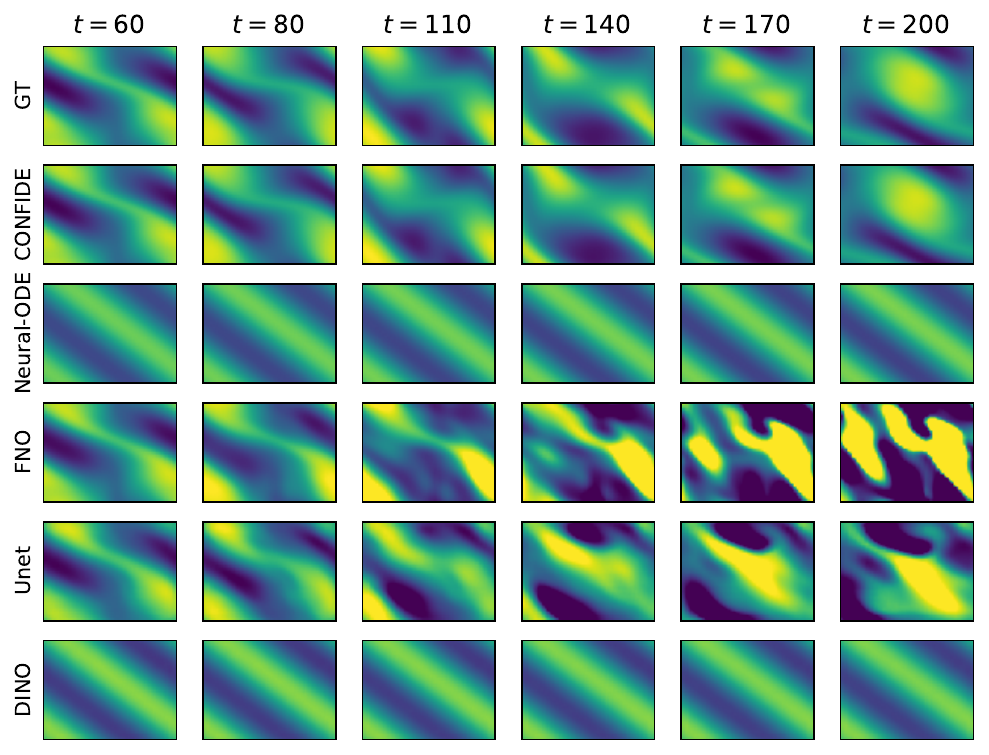}
    \caption{2D-Navier-Stokes: Figures in the top row show the ground truth, i.e., the PDE state $w$ for different time points, while the rows below show its estimation by the different approaches. All approaches observe the first 50 time points and predict the next 150.
    CONFIDE near-perfectly predicts the given signal even when the horizon increases. 
    }
    \label{fig:experiments.ns}
    \Description{Navier-Stokes example results at different time points. All approaches observe the first 50 time points and predict the next 150.
    CONFIDE near-perfectly predicts the given signal even when the horizon increases.}
\end{figure}

\begin{figure}[t]
\centering
\includegraphics[width=\columnwidth]{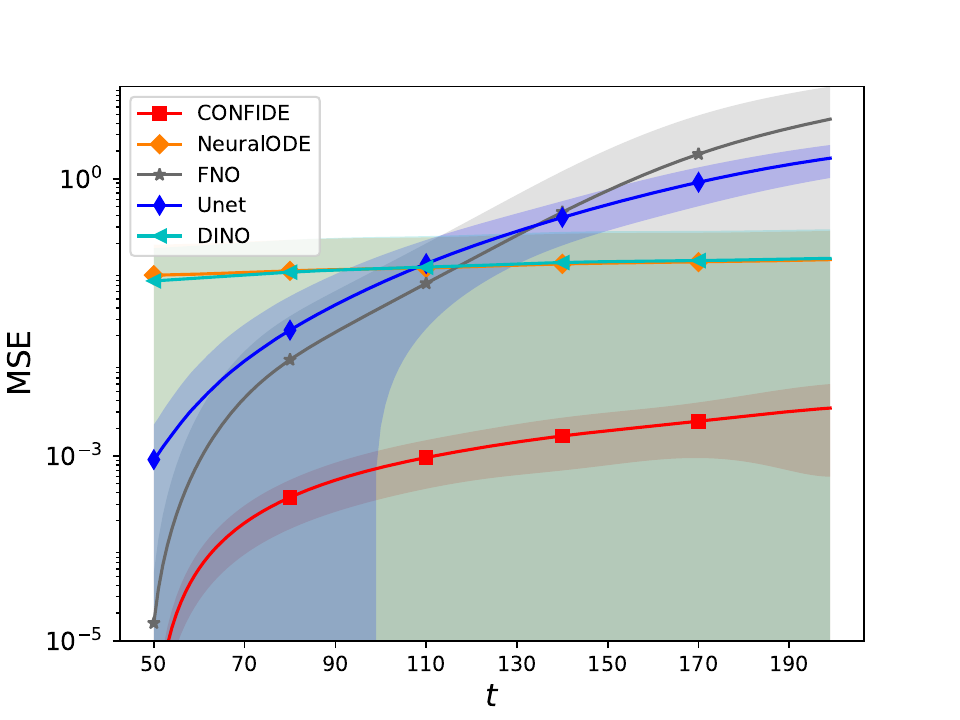}
\caption{2D-Navier-Stokes PDE: prediction error as horizon increases, for different approaches.}
\label{fig:experiments.NS_error}
\Description{Navier stokes equations - prediction error.}
\end{figure}

For the last family of PDEs, we follow \citet{yin2022continuous} and examine the Navier-Stokes equations \citep{stokes1851effect} which correspond to incompressible fluid dynamics and have the form of
\begin{align}
    \label{eq:ns}
        \fder{w}{t} = -u\nabla w + \nu \Delta u + f, \quad
        w=\nabla\times u, 
        \quad \nabla u = 0,
\end{align}
where $u$ is the velocity field, $w$ the vorticity, $\nu$ the unknown viscosity coefficient and $f$ is a constant forcing term.
The PDE state $w$ is defined over the 2-D rectangular domain $(x,y)$ with periodic boundary conditions.
The dataset created for this task consists of 1000 signals, with different values of the viscosity $\nu$.
We note that typically \citep{li2020fourier,yin2022continuous,gupta2022towards}, 
the viscosity coefficient is treated as having a single constant value among all signals in the dataset.
In this work we increase the task difficulty by creating a dataset comprised of signals that have different viscosity values.
For each signal in the dataset (both train and test) we sample a different viscosity value uniformly from $\nu\sim U[1,2]\cdot 10^{-3}$. 
We compare the prediction generated by CONFIDE to other baselines and present a typical result in Fig.~\ref{fig:experiments.ns}. In Fig.~\ref{fig:experiments.NS_error} we present the prediction error as a function of the prediction horizon. 

We summarize the results of experiments for signal prediction across all setups and approaches in Table ~\ref{tbl:exp.summary}. The table includes results for CONFIDE, all baselines, and also a variant of CONFIDE which we refer to as CONFIDE-0. This zero-knowledge variant is applicable when we know that the signal obeys some differential operator $F$, but have no details regarding the actual structure of $F$. Thus, CONFIDE-0 does not estimate the equation parameters, and only yields a prediction for the signal, utilizing our context-based architecture. We elaborate further in Appendix~\ref{app:implementation}.
We note that Neural-ODE and DINO, which are integration-based approaches, converged to a solution resembling the average of the observed signal without any dynamics evolution over time.
This issue has also been demonstrated and discussed in several other related works \citep{abrevaya2023goku,turan2021multiple,iakovlev2022latent}.

\begin{table}[ht!]
\caption{Coefficient estimation error for different experimental setups: constant coefficients, Burgers' equation, two-dimensional FitzHugh-Nagumo and two-dimensional Navier-Stokes. The variance is calculated over the entire test set, namely 1000 signals for the first two setups and 100 signals for the last two.}
\centering
        \begin{tabular}{cc}
            \toprule
            Setup & Coefficient estimation error \\
            \midrule
            Constant coeff. & $0.0095 \pm 0.0131$ \\
            Burgers' & $0.0454 \pm 0.0333$ \\
            FitzHugh-Nagumo-2D & $0.0075 \pm 0.0123$ \\
            Navier-Stokes-2D & $1.5\cdot 10^{-8} \pm 1.0\cdot 10^{-8}$ \\
            \bottomrule
        \end{tabular}
\label{tbl:experiments.all_error}
\end{table}

\begin{table*}[t]
\caption{Result summary for the signal prediction task, on all four PDE systems, together with two OOD experiments. The numbers represent signal prediction error at the end of the prediction horizon, averaged over the entire test set.}
% \vskip 0.15in
\label{tbl:exp.summary}
\begin{center}
% \begin{small}
\begin{tabular}{lcccc|cc}
\toprule
Method & \shortstack{Constant \\ coefficients} & Burgers' & \shortstack{FitzHugh-\\Nagumo} & \shortstack{Navier-\\Stokes} & \shortstack{Burgers' OOD \\ Initial conditions} & \shortstack{Burgers' OOD \\ Coefficients} \\
\midrule
% \multirow{5}{*}{\shortstack{FitzHugh-\\Nagumo}} 
CONFIDE & $0.0023 \pm 0.0036$ & $0.0008 \pm 0.0011$ & $0.0083 \pm 0.0177$ & $0.0033 \pm 0.0027$ & $0.0010 \pm 0.0012$ & $0.0074 \pm 0.0100$\\
CONFIDE-0 & $0.0079 \pm 0.0218$ & $0.0009 \pm 0.0016$ & $0.0845 \pm 0.0978$ & $0.0173 \pm 0.0355$ & $0.0020 \pm 0.0022$ & $0.0057 \pm 0.0091$\\
Neural-ODE & $0.0680 \pm 0.0905$ & $0.0272 \pm 0.0627$ & $0.2944 \pm 0.2293$ & $0.1334 \pm 0.1391$ & $0.0133 \pm 0.0208$& $0.0423 \pm 0.0649$\\
FNO & $0.0538 \pm 0.0680$ & $0.9351 \pm 0.3091$ & $2.5727 \pm 17.732$ & $4.4223 \pm 5.5352$ & $0.9367 \pm 0.2322$ & $0.9646 \pm 0.3304$\\
Unet & $0.0160 \pm 0.0199$ & $0.0016 \pm 0.0023$ & $0.1293 \pm 0.1748$ & $1.6712 \pm 0.6500$ & $0.0015 \pm 0.0024$ & $0.0096 \pm 0.0121$\\
DINO & $0.0850 \pm 0.0994$ & $0.0142 \pm 0.0206$ & $0.1651 \pm 0.1279$ & $0.1378 \pm 0.1462$ & $0.0142 \pm 0.0189$ & $0.0336 \pm 0.0334$\\
\bottomrule
\end{tabular}
% \end{small}
\end{center}
\end{table*}

\subsection{Out-Of-Distribution Data}
In this subsection we provide additional experiments conducted on out-of-distribution (OOD) data. These experiments were selected to demonstrate how CONFIDE can handle observations that are significantly different than the data in the train set.
We divide the OOD experiments into two parts: (1) the initial conditions observed are not smooth and have some discontinuity, and (2) the parameters used to generate the signals in the test set are sampled from a different distribution than the one used in the train set.

\paragraph{\textbf{Non-smooth initial conditions}}
In this first benchmark, we demonstrate how CONFIDE handles the case where the test data has a discontinuity point in the test set, while it was trained on continuous data only.
The importance of this test is mainly because CONFIDE evaluates the spatio-temporal derivatives of the signal numerically using a finite-differences approach.
This computation might result in very high derivatives in these non-smooth locations and interfere with the ability of the algorithm to provide reliable predictions.
For this task, we generated a new test set based on the Burgers' equation experiment, but the initial conditions are sampled to demonstrate discontinuity in $u(t=0,x=L/2)$.
We stress that the train-set is still the original one, since our goal is to  test whether CONFIDE is able to handle OOD data, which, in this case, comes in the form of OOD initial conditions.
As shown in Table~\ref{tbl:exp.summary}, CONFIDE's prediction error remains low, suggesting that it successfully predicts the given observations and scores close to the original score on the original burgers' test-set.

\paragraph{\textbf{OOD coefficients}}
In the second benchmark, we demonstrate how CONFIDE handles the case where the observed signal in the test set is generated from a PDE with coefficients that come from a distribution different from the ones in the train set.
For this task, we generated a new test set based on the Burgers' equation experiment, where the coefficient $a$ is sampled from $u\sim U[2,4]$ instead of $u\sim U[1,2]$ as in the train set. 
This modification in the coefficient distribution, results in generated signals that might be significantly different than the ones observed in the train set.
As shown in Table~\ref{tbl:exp.summary}, CONFIDE continues to provide good results compared to other baselines.
We note that since CONFIDE learns to output coefficients only in the range of the coefficients in the train set, it projects the observed signal to the range of coefficients in the train set so that it best describes the observed signal.

\begin{table}[H]
\centering
        \begin{tabular}{lcc}
            \toprule
            Algorithm & \shortstack{Prediction MSE \\ OOD initial conditions} & \shortstack{Prediction MSE \\ OOD coefficients}\\
            \midrule
    CONFIDE & $0.0010 \pm 0.0012$ & $0.0074 \pm 0.0100$\\
    Neural-ODE & $0.0133 \pm 0.0208$ & $0.0423 \pm 0.0649$\\
    FNO & $0.9367 \pm 0.2322$ & $0.9646 \pm 0.3304$\\
    Unet & $0.0015 \pm 0.0024$ & $0.0096 \pm 0.0121$\\
                \bottomrule
        \end{tabular}
\caption{Results summary on the two OOD benchmarks for different approaches.}
\label{tbl:app_ood}
\end{table}

\subsection{CONFIDE vs Neural ODE (Finite-Differences vs Integration)}
A key part in the CONFIDE algorithm is its use of finite-differences to evaluate the spatio-temporal derivatives.
As we demonstrate, CONFIDE can successfuly use this approach to provide both coefficient estimation and reliable predictions.
However, another important advantage that a finite-differences approach might have over ``integration'' approaches (such as the adjoint method used in Neural-ODE \citep{chen2018neural} and DINo \citep{yin2022continuous}) is the amount of time required for training the model.

To demonstrate this effect we created a toy example based on an ODE of a frictionless single pendulum: $\ddot{\theta} = - g/l \cdot \sin(\theta)$,
where $g$ is the gravitational parameter, $\theta$ is the angle of the pendulum, and $l$ is the length of the pendulum.
The two algorithms compared are CONFIDE and an ODE-aware version of Neural-ODE, where it is also given the ODE form of the pendulum.
Both algorithms are presented with the same information and have the same goals: estimating the length of the pendulum on a given signal $l$, and predicting the future of the observed signal at $t>T$.

CONFIDE evaluates the time derivative of the observed signal ${d^2\theta}/{dt^2}$ via finite-differences, and forces the derivative to be similar to the rest of the ODE. 
Neural-ODE uses the initial value ($\theta(t=0)$) and $\hat{l}$ to generate the signal $\hat{\theta}(t=0,\ldots,T)$ by integrating via an ODE-solver, and then optimizes the generated signal to match the observed one.
We trained both algorithms using the exact same setting, and observed that not only were CONFIDE's results better, but it also trained significantly faster. Specifically, CONFIDE's training time was 3.6 seconds, and Neural-ODE's training time was 159.2 seconds, making CONFIDE \textbf{$\sim 44$ times faster}.

\subsection{Autoencoder ablation study}
\label{sec:app.ablation}
\begin{figure}[ht!]
    \centering
    \includegraphics[width=\columnwidth]{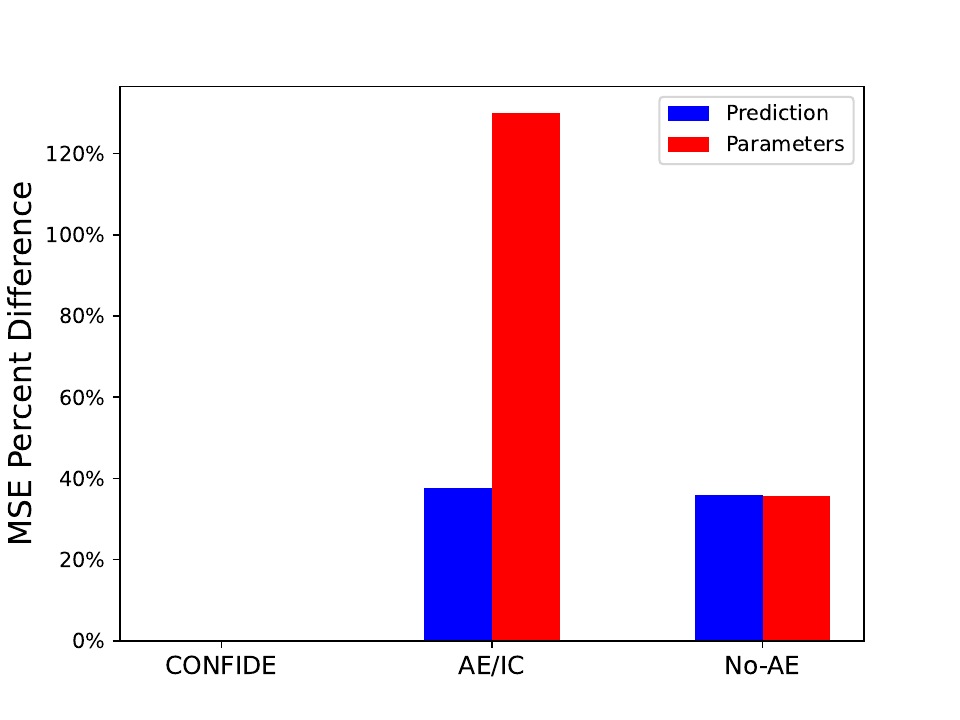}
    \caption{Ablation study on the constant coefficients PDE dataset. The Y axis shows the percentage difference between the different approaches and the standard CONFIDE one (thus it scores 0\%). We demonstrate the effects on both signal prediction (blue), and parameter estimation (red).}
    \label{fig:ae_ablation}
    \Description{Ablation study that shows that the autoencoder with initial conditions is the best architecture.}
\end{figure}

In this section, we demonstrate the effect that adding a decoder network has on CONFIDE.
To this end, we evaluate three different scenarios:
\begin{itemize}
\item \textbf{CONFIDE.} Using a decoder followed by a reconstruction loss, and feeding the initial conditions in addition to the latent vector (demonstrated in the text as initial conditions aware autoencoder)
\item \textbf{AE-IC.} Similarly, using a decoder followed by a reconstruction loss, but the autoencoder is not initial-conditions aware.
\item \textbf{No-AE.} The network trains solely on the PDE loss, without the decoder part (i.e., by setting $\alpha=0$).
\end{itemize}
Results of the three approaches on the constant PDE dataset are shown in Fig.~\ref{fig:ae_ablation}.
When comparing a setup with no decoder part (i.e., No-AE) with a setup that has a decoder, but does not use the initial conditions as a decoder input (i.e., AE/IC), we observe that merely adding a decoder network might have a negative effect on the results, especially when analyzing the parameter estimation results.
One reason for this may be that the neural network needs to compress the observed signal in a way that should both solve the PDE and reconstruct the signal. This modification of latent space has a negative effect in this case.
When also adding the initial conditions as an input to the decoder (i.e., the standard CONFIDE), we observe significant MSE improvement in both signal prediction and parameter estimation ($\sim$35\% improvement in both).
This result suggests that adding the initial conditions aware autoencoder enables the networks to learn a good representation of the dynamics of the observed signal in its latent space.

%% file: conclusions.tex
In this work we introduce a new hybrid modelling approach, combining mechanistic knowledge with data. The knowledge we assume is in the form of a PDE family, without specific coefficient values, typically supplied by field experts. 
% The dataset we rely on is readily available in physical modelling problems, as it is simply a collection of spatio-temporal signals belonging to the same PDE family, with different coefficients. 
\balance
The problem we introduce in this work is unique because the signals at inference time correspond to PDEs with coefficients that differ from those in the training data. 
This makes the prediction problem similar to a zero-shot prediction challenge, as the model needs to generalize to unseen dynamics.
% Unlike other schemes, CONFIDE does not require knowledge of the coefficients of the PDE generating our train data, but learns how to estimate them.
We conduct extensive experiments on four well-known PDE systems, comparing our scheme to other solutions and testing its performance in different regimes. 
CONFIDE outperforms all baselines, provides reliable PDE coefficient estimations, robust to different values of hyper-parameters, and scores well even when the test signals come from out-of-distribution signals.
We further stress-test CONFIDE by removing most of the mechanistic knowledge it receives (namely, CONFIDE-0, for zero knowledge), and show that it is still able to outperform other baselines.
There are many promising future directions, such as scaling CONFIDE to real-world problems like the ones mentioned in Section~\ref{sec:related}.

% Future directions we would like to pursue include a straightforward extension to handle signals with missing datapoints, handling ``out of distribution'' signals, generated by parameters beyond the support of the dataset, and examining the robustness of predicting such signals. Another question that comes to mind is whether including multiple signals generated by the same parameters has an effect on quality of results, similar to or different from that of the context ratio. Finally, we are eager to apply CONFIDE to a real world problem like the ones mentioned in Section~\ref{sec:related}.

%% file: appendix.tex
\section{Numerical scheme}\label{sec:app_num}
\input{appendix/numerical}

\section{Experimental and implementation details}\label{app:exp}
\input{appendix/experiments}

%% file: appendix/numerical.tex
The partial derivatives are estimated using standard numerical schemes for each point in time and space. We choose discretization parameters $\Delta x$ for the spatial axis and $\Delta t$ for the temporal axis where we solve the PDE numerically on the grid points $\{(i\Delta x, j\Delta t)\}_{i=0, j=0}^{N_x, N_t}$ with $L=N_x \Delta x$ and $T=N_t \Delta t$. Let us denote the numerical solution with $\hat{u}_{i,j}$. We use the \emph{forward-time central-space} scheme, so a second order scheme from \eqref{eq:general_PDE} would be
\begin{equation}
\begin{split}\label{eq:triangle_scheme}
\frac{\hat{u}_{i,j+1}-\hat{u}_{i,j}}{\Delta t}
=&
p_2(i,j,u(i,j))\frac{\hat{u}_{i+1,j}-2\hat{u}_{i,j} + \hat{u}_{i-1,j}}{\Delta x^2}\\
&+p_1(i,j,u(i,j))\frac{\hat{u}_{i+1,j}-\hat{u}_{i-1,j}}{2\Delta x}\\
&+ p_0(i,j,u(i,j))
\end{split}
\end{equation}
We refer the reader to  \cite{strikwerda2004finite} for a complete explanation. 

%% file: appendix/experiments.tex
We provide further information regarding the experiments described in Section~\ref{sec:experiments}. We ran all of the experiments on a single GPU (NVIDIA GeForce RTX 2080), and all training algorithms took $<10$ minutes to train.
All algorithms used ~5-10M parameters (more parameters on the FitzHugh-Nagumo experiment).
Full code implementation for creating the datasets and implementing CONFIDE and its baselines is available in \url{https://github.com/orilinial/CONFIDE}.

\subsection{Dataset details}
To create the dataset, we generated signals using the \verb|PyPDE| package, where each signal was generated with different initial conditions.
In addition, as discussed in Section~\ref{sec:related}, we made an important change that makes our setting much more realistic than the one used by other known methods: the PDE parametric functions (e.g., $(a,b,c)$) are sampled for each signal, instead of being fixed across the dataset, making the task much harder.
To evaluate different models on the different datasets, we divided the datasets into 80\% train set, 10\% validation set and 10\% test set. 
\begin{figure}
    \begin{center}
    \includegraphics[width=\columnwidth]{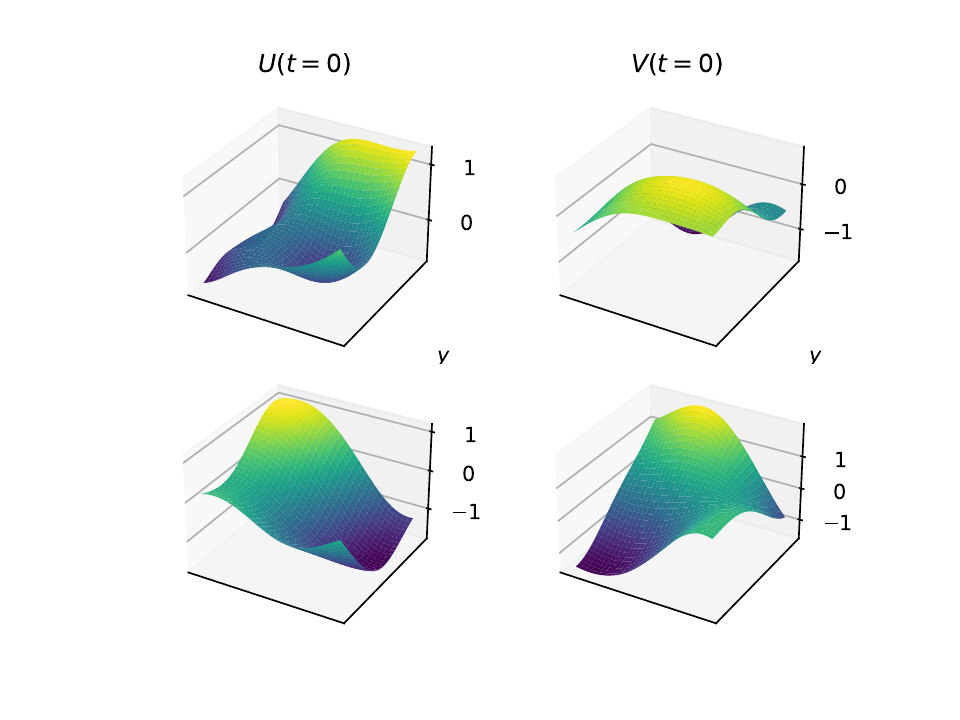}
    \end{center}
    \caption{Two examples of initial conditions for the 2D FitzHugh-Nagumo datasets. The left column describes the first state variable $u$, and the right column is the state variable $v$. Top row is the first example, and the bottom row is the second. All initial conditions are drawn from a GP prior not constrained to boundary conditions.}
    \label{app:fn2d_ic_demonstration}
    \Description{Examples of initial conditions for the FitzHugh-Nagumo dataset.}
\end{figure}

\textbf{Second Order PDE with Constant Coefficients.}
For this task, we generated 10,000 signals on the spatial grid $x\in[0,20]$ with $\Delta x=0.5$, resulting in a spatial dimension consisting of 40 points.
Each signal was generated with different initial conditions sampled from a Gaussian process posterior that obeys the Dirichlet boundary conditions~$u(x=0)=u(x=L)=0$.
The hyper-parameters we used for the GP were $l=3.0, \sigma=0.5$, which yielded a rich family of signals.
The parameter vector was sampled uniformly: $a \sim U[0,2]$, $b$ and $c \sim U[-1,1]$ for each signal, resulting in various dynamical systems in a single dataset.
To create the signal we solved the PDE numerically, using the explicit method for times $t\in[0, 5.0]$ and $\Delta t = 0.05$.
Signals that were numerically unstable were omitted and regenerated, so that the resulting dataset contains only signals that are physically feasible.

\textbf{Burgers' PDE.}
To create the Burgers' PDE dataset we followed the exact same process as with the constant coefficients PDE, except for the parameter sampling method.
Parameter $a$ was still drawn uniformly: $a \sim U[1,2]$, but $b$ here behaves as a function of $u$: $b(u) = -u$, commonly referred to as the viscous Burgers' equation.

\textbf{FitzHugh-Nagumo equations.}
For the purpose of creating a more challenging dataset with two spatial dimensions we followed \citet{yin2021augmenting}, and used the 2-D FitzHugh-Nagumo PDE (described in Eq.~\ref{eq:fn2d_eq}).
To make this task even more challenging and realistic, we created a small dataset comprising only 1000 signals defined on a 2D rectangular domain, discretized to the grid $[-0.16, 0.16] \times [-0.16,0.16]$.
The initial conditions for each signal were generated similarly to the other experiments, by sampling a Gaussian process prior with $l=0.1$, which generated a rich family of initial conditions, as can be seen in Fig.~\ref{app:fn2d_ic_demonstration}.
To create the coefficient function we sample $k\sim U[0,1]$ per signal, and set $(a,b)=(1e-3, 5e-3)$.
To create the signal we solved the PDE numerically, using the explicit method for times $t\in[0, 1.0]$ and $\Delta t = 0.01$.

\textbf{Navier-Stokes equations}
For this task we follow exactly \citet{yin2022continuous} and \citet{li2020fourier} by generating a dataset of 2-D Navier-Stokes PDE \citep{stokes1851effect}.
This dataset corresponds to the incompressible fluid dynamics, and defined by Equation~\eqref{eq:ns}.
The spatial grid used for this task is $\Omega=[-1,1]^2$, with dimensions $32\times 32$ and sample the viscosity value by $\nu\sim U[1,2]\cdot 10^{-3}$.
As with the FitzHugh-Nagumo experiment, we generated a dataset of 1000 signals, 100 of which are kept as test signals.
The value of the constant forcing term is set by:
\begin{align*}
    f(x_1,x_2)=0.1\left(\sin\left(2\pi(x_1+x_2)\right)+\cos\left(2\pi(x_1+x_2)\right)\right), \forall (x_1,x_2)\in\Omega.
\end{align*}
For the time evolution settings we used $\delta_t=0.1$ and $T=20.0$. The training horizon is set at $T=5.0$ (i.e., $\rho=0.25)$. The initial conditions are sampled as in \citet{yin2022continuous}.

\subsection{Implementation details}\label{app:implementation}
\balance

\paragraph{CONFIDE}
The CONFIDE algorithm consists of two main parts: an auto-encoder part that is used for extracting the context, and a coefficient-estimation network.

The autoencoder architecture consists of an encoder-decoder network, both implemented as MLPs with 6 layers and 256 neurons in each layer, and a ReLU activation.
For the FitzHugh-Nagumo dataset, we wrap the MLP autoencoder with convolution and deconvolution layers for the encoder and decoder respectively, in order to decrease the dimensions of the observed signal more effectively.
We note that the encoder-decoder architecture itself is not the focus of the paper.
We found that making the autoencoder initial-conditions-aware by concatenating the latent vector in the output of the encoder to the initial conditions of the signal $u(t=0)$, greatly improved results and convergence time.
The reason is that it encourages the encoder to focus on the dynamics of the observed signal, rather than the initial conditions of it. We demonstrate this effect in Section~\ref{sec:app.ablation}.

The second part, which is the coefficient estimator part, is implemented as an MLP with 5 hidden layers, each with 1024 neurons, and a ReLU activation.
The output of this coefficient-estimator network is set to be the parameters for the specific task that is being solved.
In the constant-parameters PDE, the output is a 3-dim vector $(\hat{a},\hat{b},\hat{c})$.
In the Burgers' PDE, the output is composed of a scalar $\hat{a}$, which is the coefficient of $\sder{u}{x}$ and the coefficient function $b(u)$, which is a vector approximating the coefficient of $\fder{u}{x}$ on the given grid of $x$.
In the FitzHugh-Nagumo PDE, the output is a scalar $k$ used for inferring $R_u(u,v,k)$, and the function $R_v$ on the 2D grid $(x,y)$.

The next step in the CONFIDE algorithm is to evaluate the loss which is comprised of two losses: an autoencoder reconstruction loss $\mathcal{L}_{AE}$, and a PDE functional loss $\mathcal{L}_{coef}$.
The autoencoder loss is a straightforward $L^2$ evaluation on the observed signal $u^c$ and the reconstructed signal.
The functional loss is evaluated by first numerically computing all the derivatives of the given equation on the observed signal.
Second, evaluating both sides of the differential equations using the derivatives and the model's coefficient outputs, and lastly, minimizing the difference between the sides.
For example, in the Burgers' equation, we first evaluate $\fder{u}{t}$, $\sder{u}{x}$, and $\fder{u}{x}$, we then compute the coefficients $\hat{a}$ and $\hat{b}(u)$, and finally minimize:
\begin{align*}
    \mathop{\min}\limits_{\omega}\left\Vert \fder{u}{t} - \hat{a} \cdot \sder{u}{x} - \hat{b}(u) \cdot \fder{u}{x} \right\Vert.
\end{align*}
Since this algorithm evaluates numerical derivatives of the observed signals, it could be used for equations with higher derivatives, such as the wave equation, for instance.

\paragraph{CONFIDE-0.}
Similarly to the standard CONFIDE algorithm, we consider a zero-knowledge version, where we only know that the signal obeys some differential operator $F$, but have no details regarding the actual structure of $F$. Thus, the input for the coefficient-estimator network is the current PDE state ($u$ in the 1D experiment and $(u,v)$ in the 2D experiment), and the latent vector extracted from the auto-encoder.
The model then outputs an approximation for time derivative of the PDE states,
i.e., the model's inputs are $(u_{t}, g_\phi(u^c))$ and the output is an approximation for $\fder{u}{t}$.
The optimization function for this algorithm therefore tries to minimize the difference between the numerically computed time derivative and the output of the model:
\begin{align*}
    \mathcal{L}_{\text{CONFIDE-0}} = \alpha\cdot\mathcal{L}_{AE}+(1-\alpha)\cdot \sum_{i=1}^N \left\Vert \fder{u}{t} - m_\theta(u_i^c, g_\phi (u_i^c)) \right\Vert ^2,
\end{align*}
where $\mathcal{L}_{AE}$ is defined in Eq.~\ref{eq:autoencoder_loss}, $\fder{u}{t}$ is evaluated numerically, $m_\theta$ is the network estimating the temporal derivative, $g_\phi$ is the encoder network, and $u_i^c$ is the observed signal.

\paragraph{Hyper-parameters}
For both versions of CONFIDE we used the standard Adam optimizer, with learning rate of $1e^{-3}$, and no weight decay.
For all the networks we used only linear and convolution layers, and only used the ReLU activation functions.
For the $\alpha$ parameter we used $\alpha=0.5$ for all experiments, and all algorithms, after testing only two different values: $0$ and $0.5$ and observing that using the autoencoder loss helps scoring better and faster results.

\paragraph{Neural-ODE}
We implement the Neural-ODE algorithm as suggested by \citet{chen2018neural}, section 5.1 (namely, Latent-ODE).
We first transform the observed signal through a recognition network which is a 6-layer MLP. 
We then pass the signal through an RNN network backwards in time.
The output of the RNN is then divided into a mean function, and an std function, which are used to sample a latent vector. 
The latent vector is used as initial conditions to an underlying ODE in latent space which is parameterized by a 3-layer MLP with 200 hidden units, and solved with a DOPRI-5 ODE-solver.
The output signal is then transformed through a 5-layer MLP with 1024 hidden units, and generates the result signal.
The loss function is built of two terms, a reconstruction term and a KL divergence term, which is multiplied by a $\lambda_{KL}$.
After testing several optimization schemes, including setting $\lambda_{KL}$ to the constant values $\{1, 0.1, 0.01, 0.001, 0\}$, and testing a KL-annealing scheme where $\lambda_{KL}$ changes over time, we chose $\lambda_{KL}=1e^{-2}$ as it produced the lowest reconstruction score on the validation set.
We used an Adam optimizer with $1e^{-3}$ learning rate and no weight decay.

Our implementation is based on the code in \url{https://github.com/rtqichen/torchdiffeq}.

\paragraph{FNO, Unet and DINo}
For FNO we used the standard Neural-Operator package 
\url{https://github.com/neuraloperator/neuraloperator}.
For the Unet implementation we used the implementation in \url{https://github.com/microsoft/pdearena}, and for DINo, the implementation in
\url{https://github.com/mkirchmeyer/DINo}
The input we used for these algorithms is the entire signal $u_{c}$ from time $t=0$ to $t=T-2$, and the output is a prediction of the solution at the next time point $u(t=T-1)$.
The loss is therefore an MSE reconstruction loss on $u(t=T-1)$.